\documentclass[10pt,twocolumn,letterpaper]{article}

\usepackage{iccv}
\usepackage{times}
\usepackage{epsfig}
\usepackage{graphicx}
\usepackage{amsmath}
\usepackage{amssymb}
\usepackage{float}
\usepackage{bbm}
\usepackage[ruled,vlined,noend]{algorithm2e}
\usepackage{lipsum}

\usepackage{booktabs}

\usepackage[accsupp]{axessibility}  

\iftrue     
\newcommand{\dimpp}[1]{\textcolor{blue}{[DP: #1]}}
\newcommand{\antonio}[1]{\textcolor{magenta}{[AT: #1]}}
\newcommand{\ethan}[1]{\textcolor{cyan}{[EW: #1]}}
\else
\newcommand{\dimpp}[1]{\textcolor{blue}{\noindent}}
\newcommand{\antonio}[1]{\textcolor{magenta}{\noindent}}
\newcommand{\ethan}[1]{\textcolor{orange}{\noindent}}
\fi

\newcommand{\mypar}[1]{\vspace{-0mm}\noindent\textbf{#1}}

\newcommand\blfootnote[1]{%
  \begingroup
  \renewcommand\thefootnote{}\footnote{#1}%
  \addtocounter{footnote}{-1}%
  \endgroup
}

\usepackage[pagebackref=true,breaklinks=true,letterpaper=true,colorlinks,bookmarks=false]{hyperref}

\iccvfinalcopy 

\def\iccvPaperID{10821} 

\ificcvfinal\fi

\begin{document}

\title{Scaling up instance annotation via label propagation}

\author{Dim~P.~Papadopoulos$^*$\\
MIT CSAIL\\
{\tt\small dimpapa@mit.edu}
\and
Ethan~Weber$^*$\\
MIT CSAIL\\
{\tt\small ejweber@mit.edu}
\and
Antonio~Torralba\\
MIT CSAIL\\
{\tt\small torralba@mit.edu}
}

\maketitle
\ificcvfinal\thispagestyle{empty}\fi

\begin{abstract}

Manually annotating object segmentation masks is very time-consuming.  While interactive segmentation methods offer a more efficient alternative, they become unaffordable at a large scale because the cost grows linearly with the number of annotated masks. In this paper, we propose a highly efficient annotation scheme for building large datasets with object segmentation masks. At a large scale, images contain many object instances with similar appearance. We exploit these similarities by using hierarchical clustering on mask predictions made by a segmentation model. We propose a scheme that efficiently searches through the hierarchy of clusters and selects which clusters to annotate. Humans manually verify only a few masks per cluster, and the labels are propagated to the whole cluster. Through a large-scale experiment to populate 1M unlabeled images with object segmentation masks for 80 object classes, we show that (1) we obtain 1M object segmentation masks with an total annotation time of only 290 hours; (2) we reduce annotation time by 76$\times$ compared to manual annotation; (3) the segmentation quality of our masks is on par with those from manually annotated datasets.
Code, data, and models are available online\footnote{\url{http://scaling-anno.csail.mit.edu}}.
\end{abstract}
\section{Introduction}
\blfootnote{$^*$Denotes equal contribution}%
The incredible rise in computer vision over the past decade has been propelled by the creation of datasets with multi-million annotated images such as ImageNet~\cite{russakovsky15ijcv}, Places~\cite{zhou14nips} and Kinetics~\cite{kay17arxiv}. For deep learning models to continue improving with higher capacity, there is a continual need for more training data. It has been shown that the training data must grow exponentially to see linear improvements in the models~\cite{sun17iccv,Torralba11,zhu16ijcv}.

Manual annotation for instance segmentation is expensive as it requires humans to draw a detailed outline around every object in an image~\cite{cordts16cvpr,gupta19cvpr,lin14eccv,russell08ijcv,zhou17cvpr}. For example, annotating COCO~\cite{lin14eccv} required 80 seconds per mask, an image in Cityscapes~\cite{cordts16cvpr} took 1.5 hours, and annotating ADE20K~\cite{zhou17cvpr} by a single annotator took several years. At this pace, constructing a dataset with 10M masks would require more than 200k hours and cost over \$2M~\cite{lin14eccv}.

\begin{figure}[t]
\center
\includegraphics[width=\linewidth]{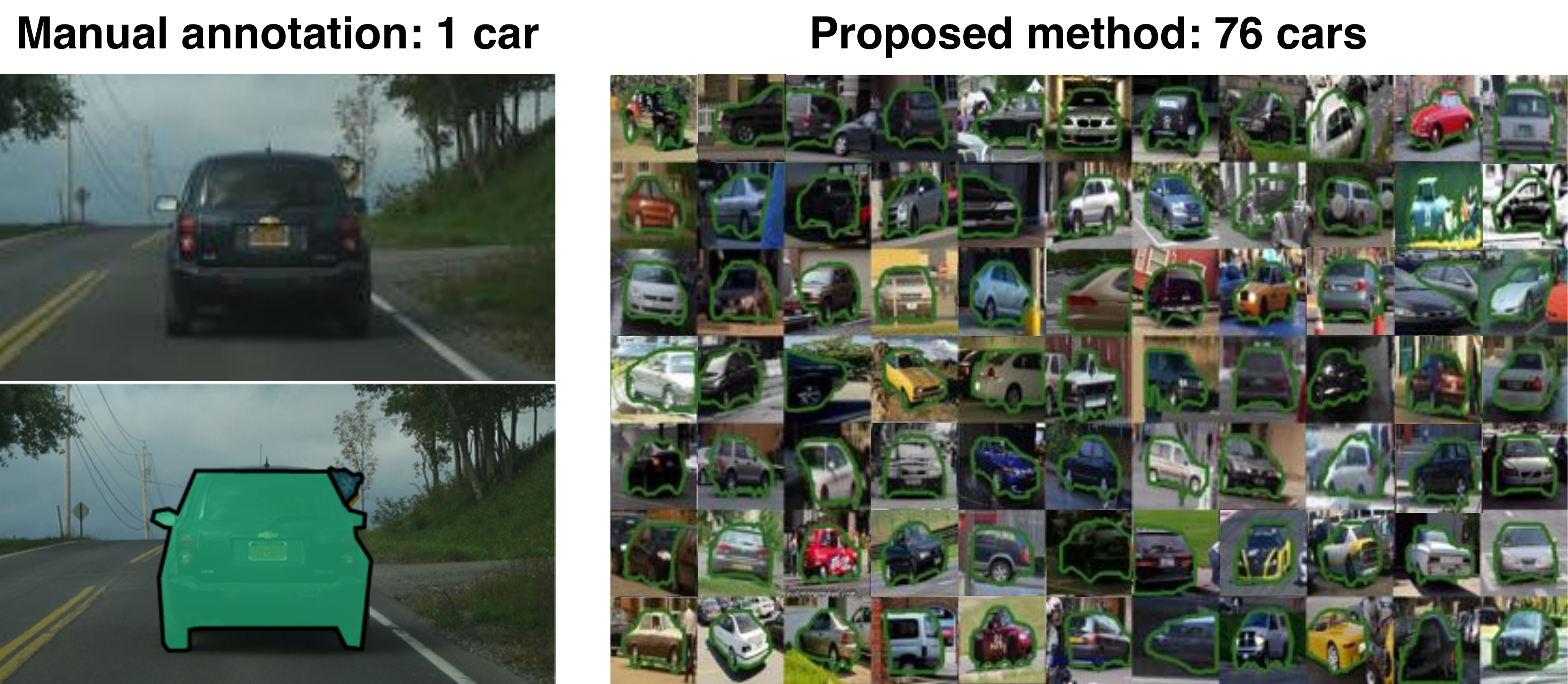}
\caption{\small \textbf{Scaling up instance annotation.} We obtain 76 segmentation masks for the cost of manually drawing one.%
}
\label{fig:teaser}
\end{figure}

An alternative is interactive segmentation~\cite{acuna18cvpr,benenson19cvpr,castrejon17cvpr,li18cvpr,ling19cvpr,papadopoulos17iccv,rother04siggraph}, where the human interaction is much faster (e.g., boxes, scribbles, clicks).
Given this interaction, these methods infer the final object mask. They offer substantial gains in annotation time and can lead to larger datasets~\cite{benenson19cvpr}.
However, because annotators intervene on every object, the cost grows linearly with the number of annotations and therefore, at a larger scale, they become unaffordable.

In this paper, we propose a method for crowd-sourcing object segmentation masks that reduces the annotation time by almost two orders of magnitude (Fig.~\ref{fig:teaser}). 
At large scale, images contain many object instances with similar appearance. We exploit this by clustering mask predictions made by an instance segmentation model. 
Human annotators \emph{verify} a few masks per cluster and we \emph{propagate} the verification labels to the whole cluster.
Since annotators interact with few masks per cluster, our cost scales with the number of clusters rather than masks.

Given a small initial image set with manually segmented objects and an annotation budget, our goal is to maximize the number of high-quality obtained annotations in an unlabeled set. We first train an instance segmentation model and obtain mask predictions on the unlabeled set and then hierarchically cluster class-specific masks.
Starting from the root of the tree, we search for candidate clusters likely to contain high-quality masks.
We ask human annotators to verify a few masks per cluster and we propagate the verification labels to the whole cluster.
Once the search terminates, the new annotations are added to the initial pool. 

We first conduct simulated experiments to explore the design space of our method. 
Then, we conduct a large-scale experiment to populate 1M unlabeled images from the Places dataset~\cite{zhou14nips} with segmentation masks for 80 object classes. 
Our results show that we obtain 1M annotations with only 290 annotation hours, reducing the annotation time by 76$\times$ compared to manual annotation~\cite{lin14eccv}. (Note that a higher annotation budget could lead to many more masks.) We also show that the mask quality is on par with those from manually annotated datasets, and our annotations leads to a better instance segmentation model than manual annotation given the same annotation time. 

\section{Related Work}
\label{sec:relwork}


\begin{figure}[t]
\center
\includegraphics[width=\linewidth]{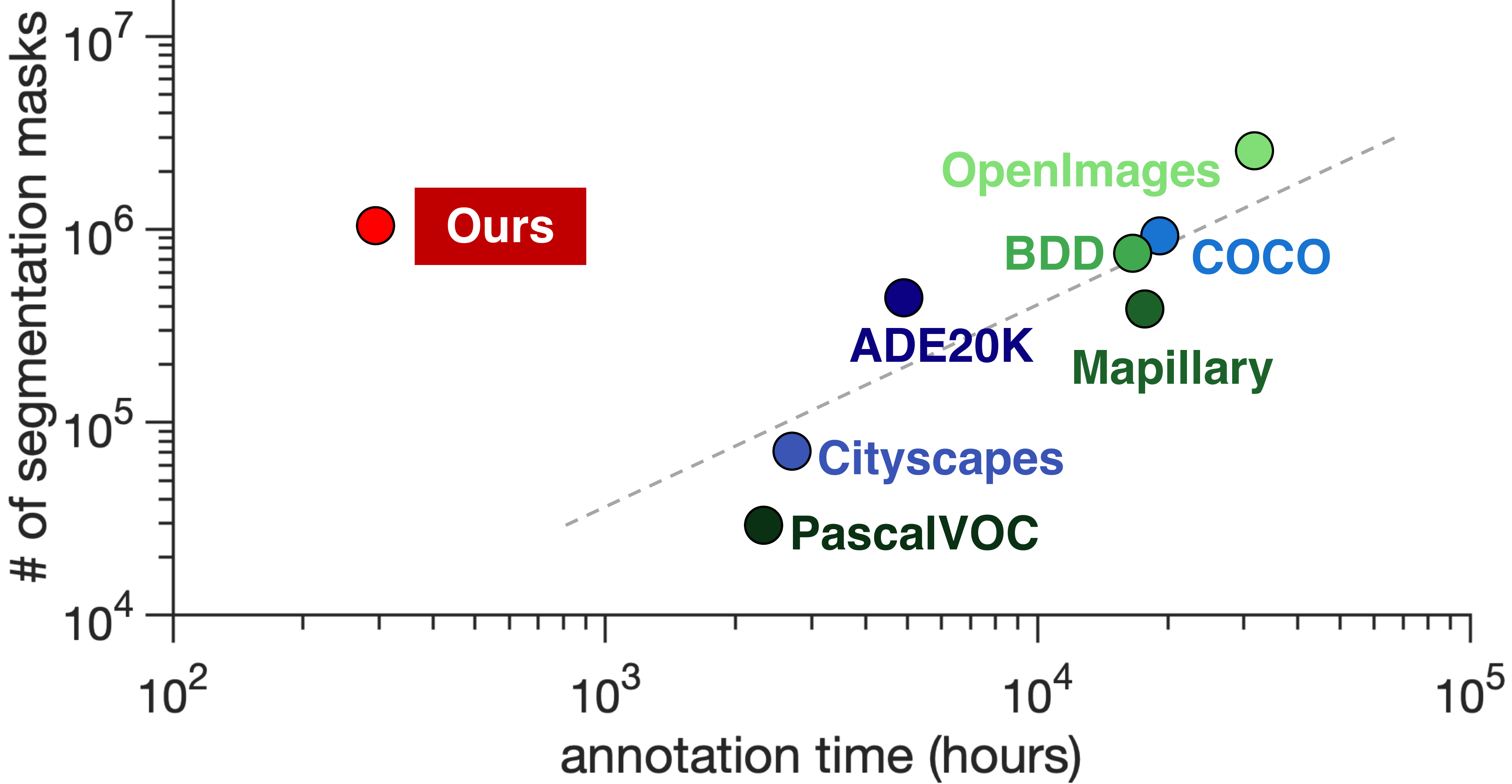}
\caption{\small \textbf{Instance segmentation datasets and annotation cost.} The size vs. the total annotation time in hours of our annotations compared to the most popular instance segmentation datasets.}
\label{fig:datasets}
\end{figure}

\mypar{Instance segmentation datasets}
are built by manually drawing accurate polygons around objects (e.g., PASCAL VOC~\cite{everingham15ijcv},
ADE20K~\cite{zhou17cvpr},
COCO~\cite{lin14eccv}, LVIS~\cite{gupta19cvpr},
Cityscapes~\cite{cordts16cvpr}, Mapillary~\cite{neuhold17iccv} and BDD100K~\cite{yu20cvpr}).
An exception is OpenImages V5~\cite{benenson19cvpr}, where the masks are collected using interactive segmentation.
Fig.~\ref{fig:datasets} shows size and estimated annotation time for each dataset. 
Increasing a dataset by an order of magnitude 
is expensive as cost grows linearly with the number of annotations (e.g., scaling COCO by 10$\times$ would require 200k hours and \$2M~\cite{lin14eccv}). 

\mypar{Annotation cost} and annotation time are proportional, so in this paper we use annotation time as it is a more objective measure than cost. The time to draw a mask or a polygon varies depending on the quality or the annotator's expertise~\cite{cordts16cvpr,everingham15ijcv,lin14eccv,yu20cvpr,zhou17cvpr}.
As a reference time for manual annotation, we use the 80 s for drawing one polygon reported in~\cite{lin14eccv}. Note that this is a conservative estimate because building a dataset requires overheads such as image classification and instance spotting~\cite{lin14eccv}.
For the total annotation time of our method, we always take into account the time of manually segmenting the initial training set and the time of verifying masks during our human-in-the-loop pipeline.

\mypar{Interactive object segmentation} 
started in the early 2000's with the seminal work of graph cut and iterative graph cuts~\cite{BoykovICCV01,BoykovPAMI04,rother04siggraph}.
After that, many approaches were proposed to reduce manual annotation effort using
bounding boxes~\cite{dai15iccv,lempitsky:iccv09,wu14cvpr}, 
point clicks~\cite{bearman16eccv,benenson19cvpr,jain16hcomp,li18cvpr,maninis18cvpr,papadopoulos17cvpr,papadopoulos17iccv,xu16cvpr}, 
scribbles~\cite{agustsson19cvpr,bearman16eccv,lin16cvpr-scribble},
and editing polygons~\cite{acuna18cvpr,castrejon17cvpr,ling19cvpr}.
These methods offer remarkable gains in time compared to manual annotation. However, because they all require human intervention for every object, the annotation cost still grows linearly with the number of object segmentation masks.

\begin{figure*}[ht!]
\center
\includegraphics[width=\linewidth]{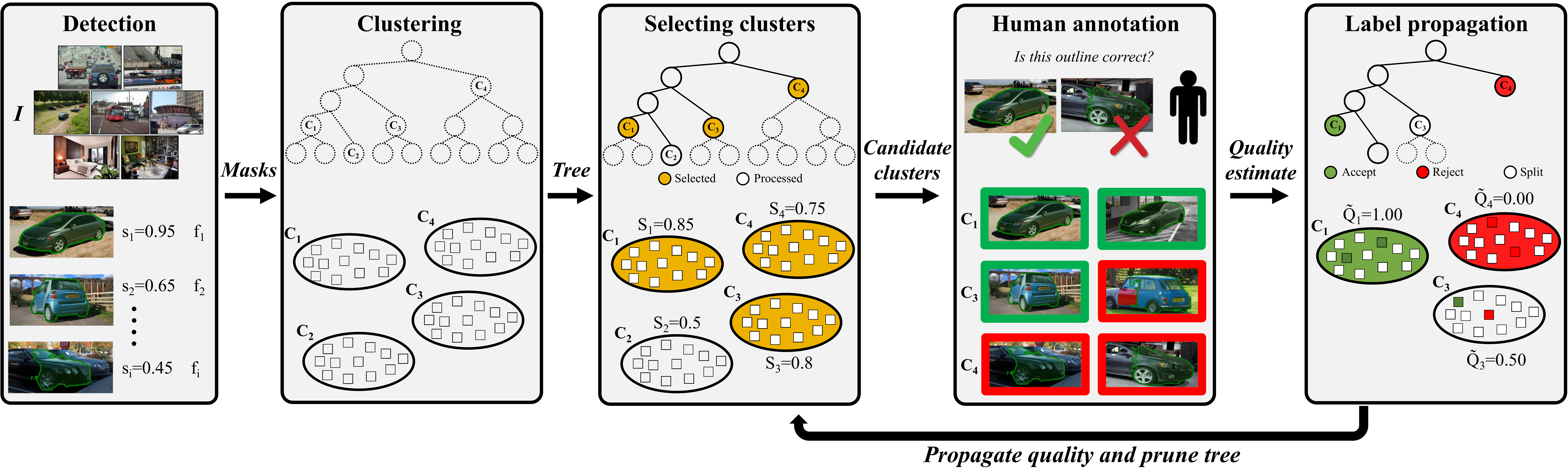} 
\caption{\small \textbf{The proposed human-in-the-loop pipeline.} We apply an instance segmentation model on an unlabeled set of images to detect object segmentation masks (\textbf{\textit{Detection}}). Then, we use hierarchical clustering to cluster masks and form a tree of clusters (\textbf{\textit{Clustering}}). We efficiently search the tree and select candidate clusters that are likely to contain high-quality masks (\textbf{\textit{Selecting clusters}}). Human annotators verify a few masks per cluster (\textbf{\textit{Human annotation}}) and we propagate the verification labels to the whole cluster (\textbf{\textit{Propagation}}). }
\label{fig:frameworkDetail}
\end{figure*}

\mypar{Annotation propagation}
uses pre-segmented images to guide the segmentation of new ones by propagating the annotation signal~\cite{guillaumin14ijcv,jain16cvpr,kim16siggraph,kuettel12eccv,rakelly18arxiv,rubinstein12eccv}. \cite{jain16cvpr} selects which images should be annotated or used for propagation, but it focuses on large objects and the task of foreground-background segmentation. \cite{kim16siggraph} uses human verification to propagate 3D shape part annotations on synthetic shape models. Here we present an active framework for instance annotation and propagation of large-scale image collections, where images contain multiple objects per image.

\mypar{Group-based labeling}  
assigns a single labeled image to a whole group
~\cite{huang11cvpr,tian07cvpr,wigness15cvpr}.
In~\cite{yu15lsun}, human annotators label only a few images, which are used to train binary classifiers and automatically label many images at once with high confidence score. In Sec.~\ref{sec:simulationsPLACES} we modify~\cite{yu15lsun} to work for instance segmentation and compare it with our method.

\mypar{Active learning}
has been used in computer vision for the tasks of image classification~\cite{joshi09cvpr,kapoor07iccv,kovashka11iccv,mac14cvpr,sinha19iccv,yoo19cvpr}, object detection~\cite{vijayanarasimhan14ijcv,yao2012cvpr} and semantic segmentation~\cite{jain16cvpr,siddiquie10cvpr,Vijayanarasimhan08nips,vijayanarasimhan09cvpr}.
The main goal of active learning is to maximize the model's performance on a test set while minimizing the number of provided human labels. Related to this goal, we aim to maximize the number of obtained annotations while minimizing the human annotation effort.

\section{Overview}
\label{sec:method_overview}

Given a small set of manually annotated images with segmentation masks and a large set of unlabeled images, our goal is to populate the unlabeled set with high-quality masks using as little human intervention as possible.
Our pipeline consists of five steps (Fig.~\ref{fig:frameworkDetail}):
(a) Detection: we deploy an instance segmentation model on an unlabeled set to obtain segmentation masks.
(b) Clustering: class-specific masks are hierarchically clustered to obtain a tree.
(c) Selecting clusters: we efficiently search the tree and select candidate clusters that are likely to contain high-quality masks.
(d) Human annotation: for each candidate cluster, we sample a few masks and ask annotators to verify whether they are correct or not. 
We use these labels to estimate the quality of the clusters.
%
(e) Propagation: If the estimated quality is very high or very low (i.e, the cluster almost exclusively contains correct or wrong masks), we propagate the verification labels to the whole cluster and set it as a leaf. Otherwise, we further split it.
%
We repeat (c), (d) and (e) to discover high-quality clusters with as few questions as possible. 

\section{Method}
\label{sec:method}

In this section, we present our pipeline to efficiently populate unlabeled images with object segmentation masks. 

\mypar{Detection.}
In this step, we train an instance segmentation model on a small initial set of manually segmented images to obtain object segmentation masks $M$ on the unlabeled set of images $I$.
For the instance segmentation model, we use PointRend~\cite{kirillov20cvpr}. Inspired by~\cite{huang19cvpr}, we modify PointRend by adding a head trained to regress the Intersection-over-Union (IoU) of the output mask using the relaxed IoU loss~\cite{li18cvpr}. We multiply this IoU score with the classification score of the box head to obtain a predicted score $s$ for every predicted mask $m$. 
Therefore, for every $m$, we get a score $s$ and a deep feature representation $f$. $f$ and $s$ together encode both the visual appearance and the mask quality.

\begin{figure}[t!]
\center
\includegraphics[width=\linewidth]{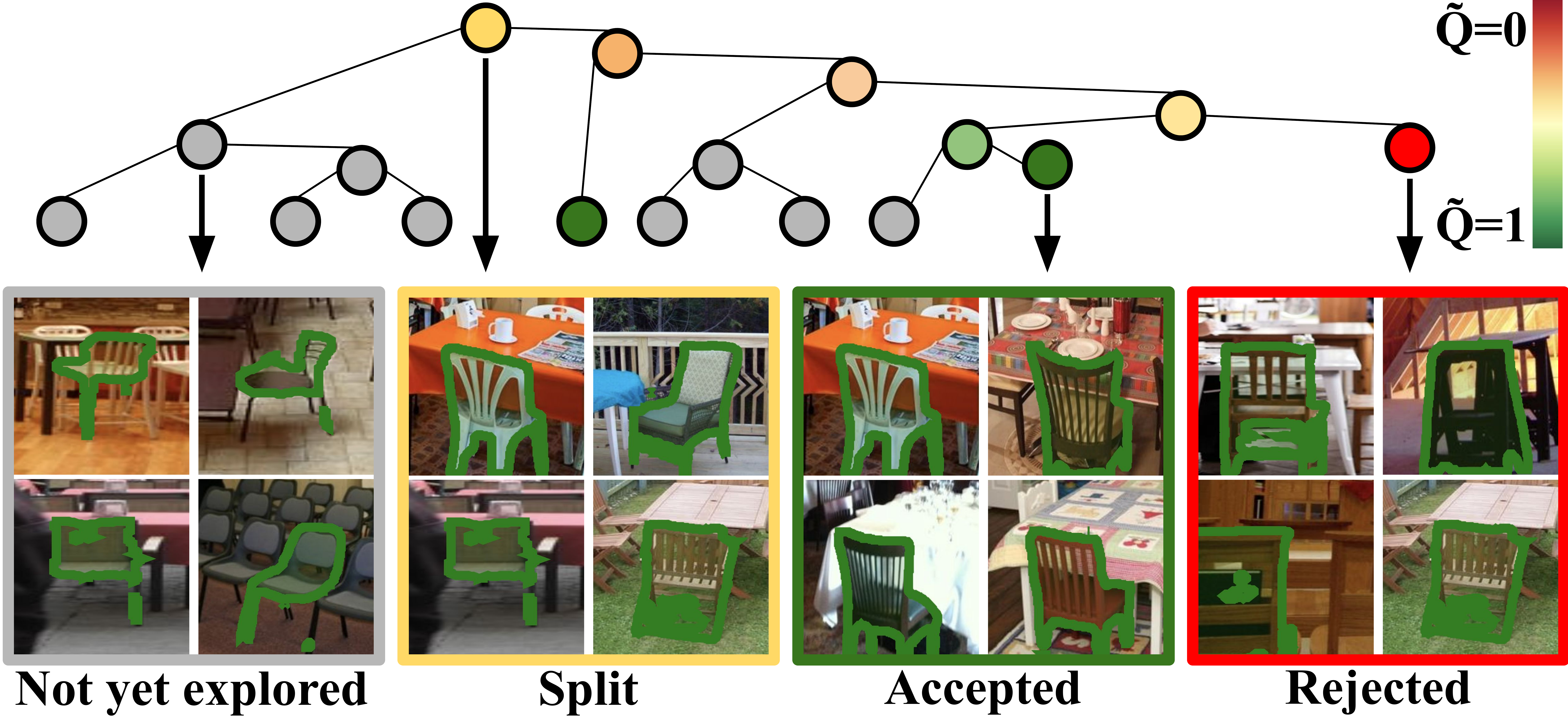}
\caption{\small \textbf{Examples of clusters in tree of the chair class.} Clusters are \emph{split} when containing both low and high quality masks, \emph{accepted} when mostly correct, and \emph{rejected} otherwise. The gray clusters have yet to be explored and have no quality estimate $\Tilde{Q}$. }
\label{fig:tree}
\end{figure}

\mypar{Clustering.}
In this step, we cluster $M$ using a feature representation $\mathcal{F}$. In Sec.~\ref{sec:simulationsADE}, we evaluate different choices for $\mathcal{F}$. For each object class, we use bottom-up hierarchical agglomerative clustering (HAC) with the complete linkage function to obtain a binary tree $T$ with clusters $C$ as vertices. A cluster $C_{j}$ is the set of vertices formed by the subtree with parent node $C_{j}$. 
The root of the tree $C_{r}$ contains all masks $M_{r}$ while the leaves are singleton clusters, i.e., $C_{j} = \{m_{j}\}$ for $j \in [1, |M_{r}|]$.
After obtaining $T$, clusters are typically defined by setting a universal threshold at a certain height in the tree. However, the clusters at the same height are not equally pure (Fig.~\ref{fig:tree}).
For this reason, in the following steps we efficiently search the tree, actively select clusters, and query annotators to find the final cluster set. 

\mypar{Selecting clusters.}
In this step, we search $T$ starting from the root node $C_{r}$ to efficiently discover a set of high-quality clusters $C^{a}$.
We use a priority queue to guide the search algorithm towards clusters that are likely to be of high quality. 
Different tree search algorithms lead to different cluster orderings, which results in a trade-off between cost and number of masks obtained during searching. However, all algorithms eventually visit the same clusters by the end of the procedure.
The majority of the clusters in $T$ are not high-quality and blindly annotating everything leads to an inferior trade-off, so instead we actively select which clusters to annotate. To do this, we estimate the likelihood of a cluster to be high-quality given the scores of the masks it contains.

A mask $m_{i}$ is considered correct when its IoU with a perfect mask $m_{i}^{*}$ capturing the object is greater than $K_{iou}$.
We define the quality of $m_i$ as $q_{i} = \mathbbm{1}_{\textrm{IoU}(m_{i}) \ge K_{iou}}$. The quality $Q_{j}$ of $C_{j}$ is defined as $Q_{j} = \frac{1}{|C_{j}|}\sum_{q_{i} \in C_{j}} q_{i}$. We consider $C_j$ to be of high quality when $Q_j \ge K_{a}$ and of low quality when $Q_j \le 1-K_{a}$.

\begin{figure}[t!]
\center
\includegraphics[width=1\linewidth]{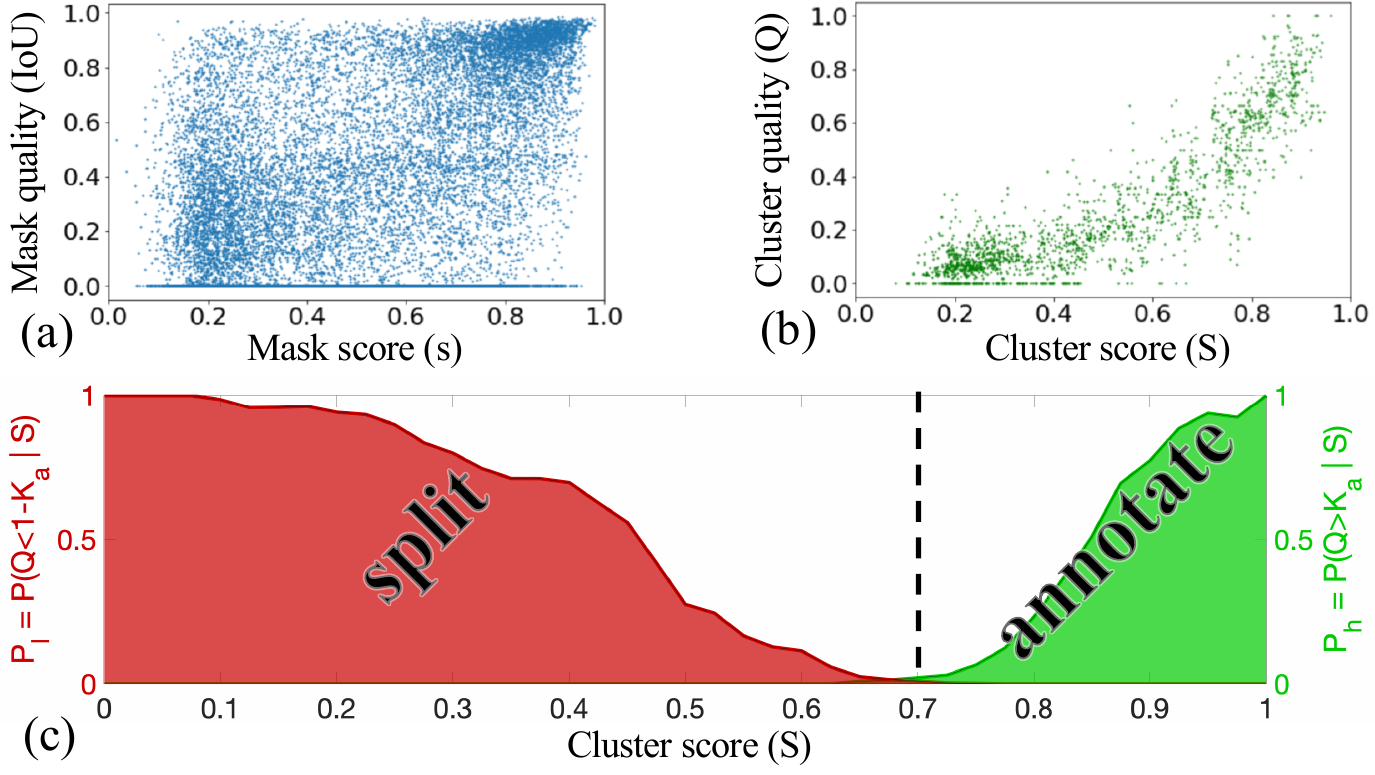}
\caption{\small \textbf{Selecting candidate clusters.} (a) The real mask IoU as a function of the mask score $s$. (b) The real cluster quality ($Q$) as a function of the cluster score $S$. We observe much stronger correlation at the cluster level.
(c) The estimated likelihood probabilities $P_h$ and $P_l$ given $S$. We only annotate clusters when $P_h > 0$. Otherwise, the cluster is split without any human intervention.}
\label{fig:activeSelection}
\end{figure}

We also define the score $S_j$ of cluster $C_j$ as the mean of the scores $s$ of the masks it contains. Therefore, the likelihood of $C_j$ to be of high quality given $S_j$ is given by $P_{h} = P(Q_{j} \ge K_{a} | S_{j})$. Similarly, $P_{l} = P(Q_{j} \le 1-K_{a} | S_{j})$ is the likelihood that $C_j$ is of low quality.
We use $S_j$ to learn the likelihoods $P_{h}$ and $P_{l}$ on a held-out set of images with ground-truths.
We denote a threshold $K_{pa}$ on $S_{j}$ such that 
$P(Q_{j} \ge K_{a} | S_{j} < K_{pa}) \approx 0$. 
When $S_{j} > K_{pa}$, the cluster is annotated, otherwise it is unlikely to be of high quality and is split to its children (Fig.~\ref{fig:activeSelection}). Alg.~\ref{algo:procedure} outlines this procedure.

\begin{algorithm}[t]
\footnotesize{
\DontPrintSemicolon
\SetAlgoLined
\KwIn{Binary tree $T$ with clusters $C$}
\KwOut{Accepted clusters $C^{a} = \{C_{j} | \Tilde{Q}_{j} \ge K_{a}\}$}
 Accepted clusters $C^{a} = \{\}$\ , Candidate clusters $C^{c} = \{C_{r}\}$\;
 \While{$|C^{c}| > 0$} {
  Sort $C^{c}$ by scores $S$\ and Pop $C_{j}$ from $C^{c}$\;
  Check $S_{j}$ to possibly split early\;
  Estimate quality $\Tilde{Q_{j}}$ of $C_{j}$ \tcp{annotate}
  \uIf{$\Tilde{Q_{j}} \ge K_{a}$} {
    $C^{a} = C^{a} \cup C_{j}$ \tcp{accept}
  }
  \uElseIf{$\Tilde{Q_{j}} \le 1 - K_{a}$} { \tcp{reject} }
  \Else {
    Children $\{C_{left}, C_{right}\} = C_{j}$\ \tcp{split}
    $C^{c} = C^{c} \cup \{C_{left}, C_{right}\}$
  }
 }
\Return $C^{a}$\;
\caption{Selecting and annotating clusters.}
}
\label{algo:procedure}
\end{algorithm}

\mypar{Human annotation.}
In this step, we ask human annotators to quickly verify whether or not a displayed mask on an object is correct or not.
This verification step is commonly used in efficient annotation schemes as each question takes less than 2 s~\cite{kuznetsova18arxiv,papadopoulos16cvpr}. Our key difference is that for each candidate cluster the annotator only verifies $N_s$ randomly sampled masks and not all of them. 
The annotators are instructed to respond positively ($l_i=1$) if the mask $m_i$ correctly outlines the target object, and negatively otherwise ($l_i=0$).
We use the human responses of the $N_s$ sampled masks $\hat{M}$ to obtain an estimated quality $\Tilde{Q}$ for a cluster as $\Tilde{Q} = \frac{1}{N_s} \sum_{l \in \hat{M}} l$.

\mypar{Label propagation.}
In this step, we use $\Tilde{Q}$ to propagate the human verification labels $l$ to the rest of the masks in the cluster. 
If $\Tilde{Q_j} \ge K_a$,  we \emph{accept} the cluster and propagate the positive verification labels ($l_i=1$) to all the masks it contains.
Similarly if $\Tilde{Q_j} \le 1-K_a$, we \emph{reject} the cluster and we propagate the negative verification labels ($l_i=0$) to all the masks it contains.
Otherwise, we further \emph{split} the cluster into its children.
The accepted and rejected clusters become leaf clusters in $T$ by pruning all their children.
The masks from the accepted clusters $C^a$ form the output annotations, while the masks from the rejected clusters are discarded.




\section{Simulated experiments}
\label{sec:simulations}

In this section, we perform simulated experiments to explore different design choices for each step of our pipeline and find parameters to minimize the annotation cost given a desired annotation quality. 
In Sec.~\ref{sec:simulationsADE}, we perform experiments on ADE20K~\cite{zhou17cvpr} while
in Sec.~\ref{sec:simulationsPLACES} we simulate a large-scale scenario to estimate parameters not possible to estimate within a small dataset. In Sec~\ref{sec:simulationsCOCO}, we perform experiments on COCO~\cite{lin14eccv} and OpenImages~\cite{benenson19cvpr}.

\mypar{Implementation details.}
For all experiments, we use PointRend~\cite{kirillov20cvpr} with FPN~\cite{lin17cvpr} and ResNet50~\cite{he16cvpr} as our instance segmentation model. During inference, we keep all predictions with a confidence score above 0.2.

\subsection{Simulated experiments on ADE}
\label{sec:simulationsADE}

\paragraph{Dataset.}
We use the training set of ADE20K~\cite{zhou17cvpr}, which consists of 20,210 images and ground-truth annotations for 2,693 classes. We select the 80 most frequent foreground object (``thing'') classes from the scene parsing ADE20K Benchmark~\cite{zhou17cvpr}. For most experiments in this section, we use 10,105 images to train the instance segmentation model and consider the other half as the unlabeled pool. We study the effect of smaller initial sets an the end of this section.

\mypar{Evaluation protocol.}
Our goal is to populate unlabeled images with high quality masks while minimizing human annotation effort. For this reason, we evaluate the trade-off between the quantity and quality of the obtained annotations versus the annotation cost. 
We measure \emph{annotation quantity} as the total number of the obtained annotations.
We measure \emph{annotation quality} as the mean IoU of the obtained annotations with respect to the ground-truth ones. 
%
We compute each metric per class and report the sum for quantity and mean for quality over all classes.
We measure \emph{annotation cost} as the total number of annotated clusters.
%
For all experiments in this section, we factor out the sampling step and when we annotate a cluster, we assume $\tilde{Q}=Q$. We consider masks with IoU $\ge 0.75$ with the ground-truth as correct and we set $K_{a} = 0.85$. 

\begin{figure*}[ht]
\center
\includegraphics[width=1\linewidth]{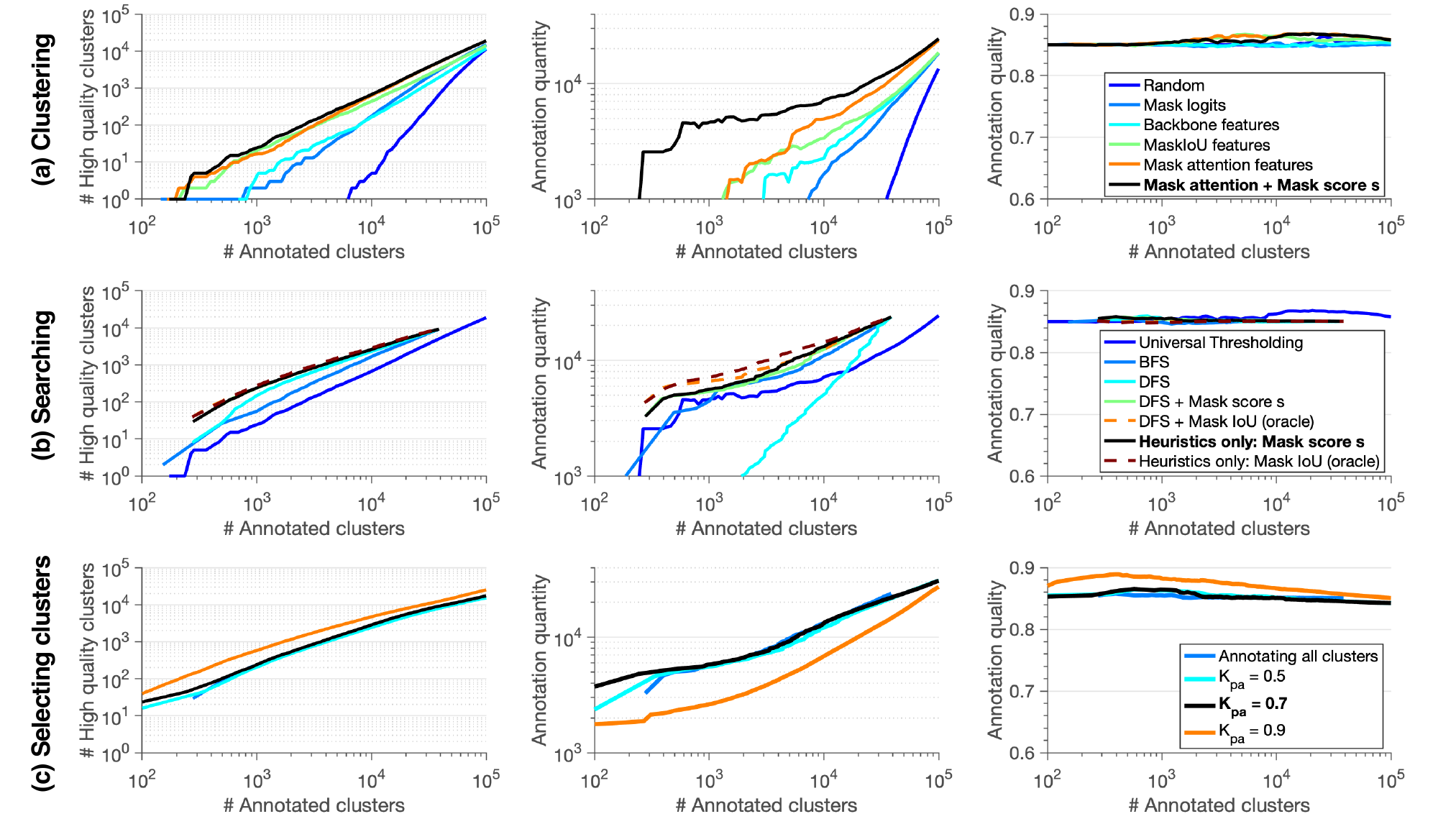}
\caption{\small \textbf{Experimental results on ADE20K.} We show the number of high-quality clusters, the annotation quantity and the annotation quality versus the number of annotated clusters in log-log scale. \textbf{(a)~Clustering}: The effect of using different feature representations $\mathcal{F} $ to construct $T$.
\textbf{(b)~Searching}: The effect of using different search algorithms when searching the tree $T$.
\textbf{(c)~Selecting clusters}: The effect of actively selecting clusters to annotate using different $K_{pa}$ values. The black lines correspond to the best result in each row.}
\label{fig:mainResultsFigure}
\end{figure*}

\mypar{Clustering.}
We first evaluate the result of the clustering step by examining how well different $\mathcal{F}$ construct $T$ (Fig.~\ref{fig:mainResultsFigure} (a)). We examine four feature types.
(a) The \emph{Mask logits} of PointRend after passing them through a sigmoid;
(b) The \emph{Backbone features} of PointRend after ROI align;
(c) The \emph{MaskIoU features} from the second last layer of the mask IoU head.
(d) The \emph{Mask attention features} where we multiply the sigmoided mask logits with the backbone features. We obtain different numbers of clusters from each $T$ by cutting at different distance thresholds in the HAC dendogram.
In Fig.~\ref{fig:mainResultsFigure}(a), we observe that the mask attention (orange line) achieves a better trade-off than the other feature types.
Combining it with the mask score $s$ (black line) leads to better clustering (more annotations given the same number of clusters); we use this in the remainder of the paper.
We also notice that the annotation quality (Fig.~\ref{fig:mainResultsFigure}(a)) is always greater than $0.85$. Even with random clustering, our verification and propagation guarantee high quality.


\mypar{Searching the tree.}
We evaluate the \emph{universal} thresholding under different number of clusters and four main search tree algorithms to search and annotate the tree:
(1) \emph{BFS}: breadth-first search;
(2) \emph{DFS}: depth-first search;
(3) \emph{DFS with heuristics}: depth-first search with a heuristic score to prioritize child clusters that are more likely to be of high quality;
(4) \emph{Heuristics only}: sort all candidate clusters according to the heuristic in decreasing order while searching the tree.
For a fair comparison, we factor out the cluster selection and we select all the clusters for annotation.

Most of the search algorithms achieve a much better trade-off than the universal thresholding (blue line) (Fig.~\ref{fig:mainResultsFigure}(b)).
\emph{Heuristics only: Mask score} achieves the best trade off (black line) and \emph{DFS} alone achieves the worst as it blindly explores a path from the root until a random leaf (cyan line). Interestingly, using the mask score $s$ as a heuristic metric we perform only slightly worse than the upper bound of using the real mean IoU of each cluster (dashed lines). We use the \emph{Heuristics only: Mask score} as our searching algorithm for the remainder of this work.

\mypar{Selecting clusters.}
The previous experiments assume every cluster is selected for verification. Here we evaluate the active selection of these clusters by following our selection mechanism described in Sec.~\ref{sec:method} using different values for $K_{pa}$. In Fig.~\ref{fig:mainResultsFigure}(c), we observe that very high values of $K_{pa}$ result in a low number of good annotations (orange line), while low values have only a very small effect to the performance (cyan line). For the remainder of this work, we set $K_{pa}=0.7$, which achieves the best trade-off (black line).

\begin{figure}[t]
\center
\includegraphics[width=\linewidth]{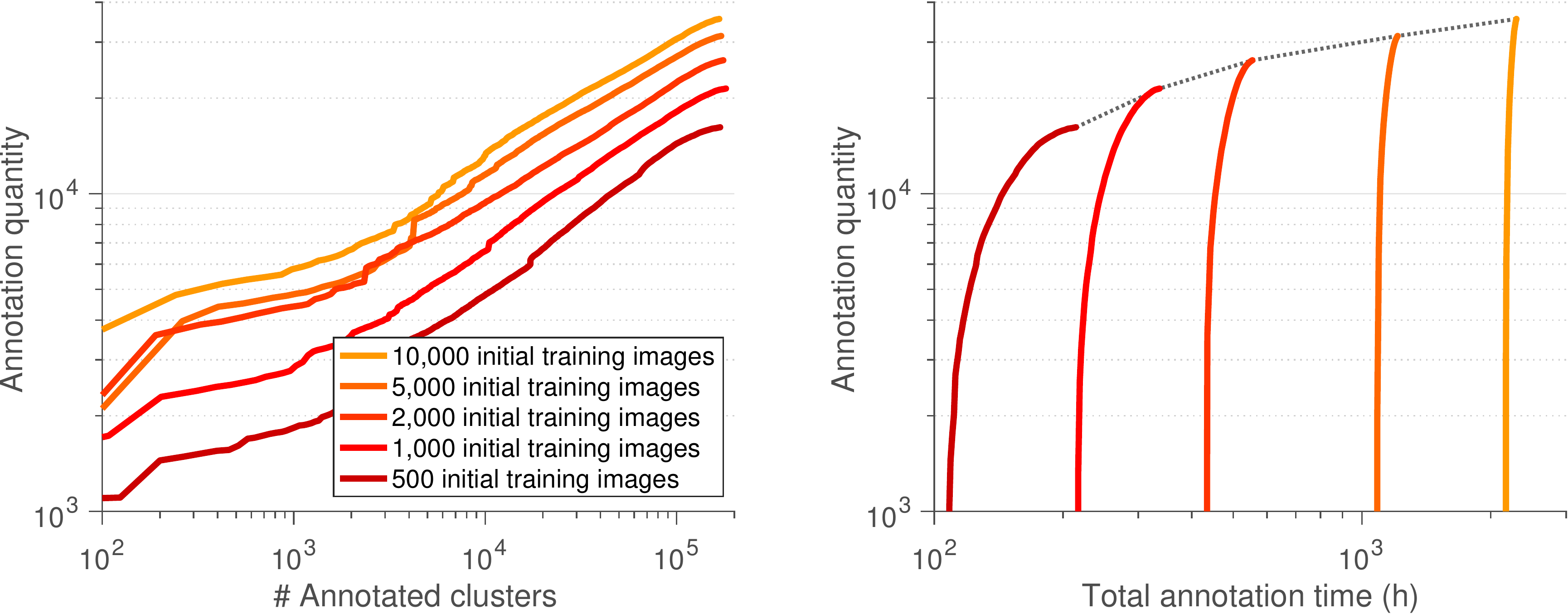}
\caption{\small \textbf{Initial training set.} (Left) The annotation quantity vs. the annotated clusters for four initial annotated sets. (Right) The annotation quantity vs. the total annotation time in hours by taking into account the time to annotate the initial set.}
\label{fig:initset}
\end{figure}

\mypar{Initial training set.}
Here we evaluate the effect of the size of the initial training set. In Fig.~\ref{fig:initset}(left), we show the annotation quantity versus the number of annotated clusters using 500, 1,000, 5,000 and 10,000 images as our initial set. We observe that with a larger set, accepted clusters contain more masks because the initial models are better. However, even with a very small set (e.g. 500 images), we obtain a significant boost in annotation quantity compared to manual annotation. Fig.~\ref{fig:initset}(right) shows the annotation quantity versus the total time including the time to manually annotate the initial set. We assume 80 s to annotate each object in the initial set and 30 s to verify each cluster. Given a fixed budget, one should select the training set that reaches the corresponding vertical point of the dotted line to maximize the annotation quantity. We use 1,000 images as our initial set for the remainder of this work as it provides a good compromise between cost and maximum annotation quantity.

\mypar{Comparison to confidence-based baseline.}
We compare our annotation quality and quantity against a simple baseline that selects masks based only on the PointRend confidence score. Our pipeline with a small annotation cost obtains 15,628 annotations with 0.83 quality. Using only the confidence score, we obtain only the top 67 high-confident annotations in order to maintain the same quality. From another perspective, the quality of the top 15,628 predictions is only 0.66. This shows that our method is doing much more than simply filtering out low confidence predictions.

\subsection{Large-scale simulated experiments}
\label{sec:simulationsPLACES}

In this section, we simulate a realistic large-scale scenario to estimate the parameters for the annotation and the propagation steps which cannot be optimized directly in ADE20K due to its small size. We estimate the number of verified samples per cluster ${N_s}$, the cluster quality threshold $K_a$, and the mask IoU threshold $K_{iou}$.  

\mypar{Evaluation set.}
The number of ground-truth masks per class in ADE20K varies from a few thousand to {\em only} a few hundred, and our experiments in Sec.~\ref{sec:simulationsADE} show that for most classes, the highly pure clusters contain on average less than five examples. 
For this reason, we design a simulated scenario under the assumption that at a larger scale the
detection masks follow a similar joint distribution $P(f, s, IoU(m))$ (Recall that $f$ is a feature representation vector, while $s$ and $IoU(m)$ are scalars). %
We learn $P$ on ADE20K in a class-specific way using a Gaussian Mixture model. Then, we sample triplets $\{f_i, s_i, IoU(m_i)\}$ and form simulated class-specific sets by following the class distribution and the number of instances per image of ADE20K. This leads to 100M triplets to run our pipeline.

\mypar{Annotation and propagation.}
Following human annotation consistency~\cite{benenson19cvpr,gupta19cvpr,kirillov19cvpr,zhou17cvpr}, a high quality instance segmentation dataset typically has a segmentation quality $SQ$ between 0.8 and 0.85. $SQ$ is defined as the average IoU for all matched manual annotations. 
We want to find the optimal values for ${N_s}$, $K_a$ and $K_{iou}$ while keeping $SQ \ge 0.85$. 
The higher the $N_s$, the more accurate the $\Tilde{Q}$ is and the less noise the final dataset contains. However, $N_s$ directly affects the annotation cost.
Setting $K_a$ or $K_{iou}$ to 1 would lead to high quality; this, however, would result to a very low number of annotations.
We run our pipeline for different parameter choices and aim to keep $SQ \ge 0.85$. We first minimize ${N_s}$ and then search over combinations of $K_a$ and $K_{iou}$ to find which values lead to the largest number of annotations. This results in ${N_s}=15$, $K_a=0.85$, and $K_{iou}=0.75$. We note that if ${N_s}=\infty$, then $SQ=0.89$.

\begin{figure}[t!]
\center
\includegraphics[width=\linewidth]{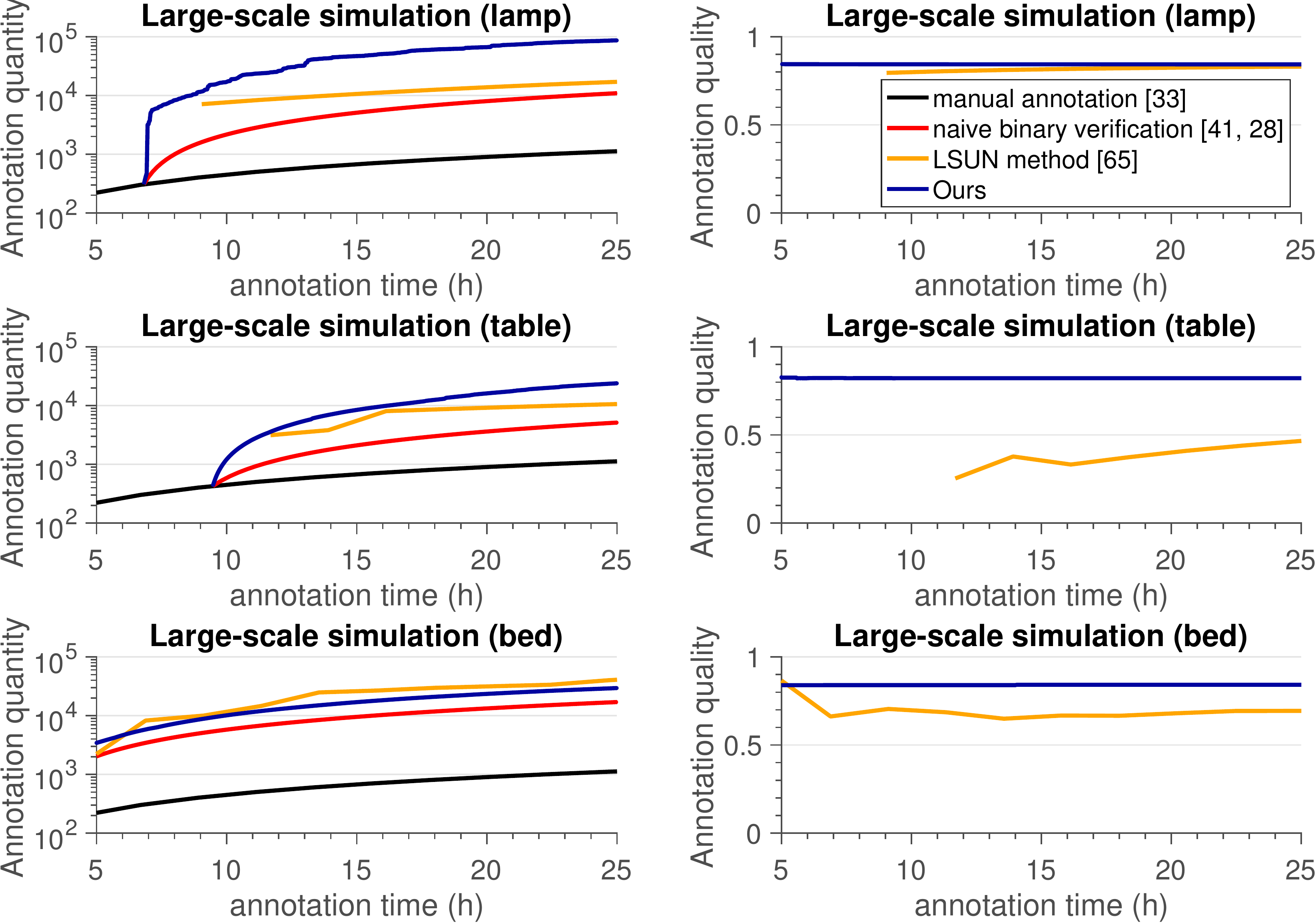}
\caption{\small \textbf{Large-scale simulated experiment.} The annotation quantity and quality vs. the annotation time in hours for 3 object classes. We compare our pipeline with manual annotation~\cite{lin14eccv}, naive binary verification~\cite{kuznetsova18arxiv, papadopoulos16cvpr} and the LSUN method~\cite{yu15lsun}.}
\label{fig:comparelsun}
\end{figure}

\mypar{Comparison to other verification approaches.}
We compare our pipeline to other verification-based approaches. In Fig.~\ref{fig:comparelsun}, we report the trade-off between annotation quantity and quality vs. the time in hours assuming 2 s for each binary question and 80 s for each manual drawing~\cite{lin14eccv}.
Our framework leads to a speed up of one to two orders of magnitude comparing to~\cite{lin14eccv}, while maintaining a high quality. %
We modified the method used for LSUN for instance segmentation~\cite{yu15lsun}, which iteratively (a) verifies a few images, (b) trains a binary classifier, and (c) applies it on unlabeled images to propagate labels and select ambiguous examples to annotate. Instead of an image pool, we use the pool of predicted masks from our initial model and apply the pipeline of \cite{yu15lsun} at masks without further changes. Our method achieves a better trade-off while maintaining a much better quality. %
Even in the object class bed, where our pipeline achieves the smallest improvement over brute-force verification~\cite{papadopoulos16cvpr}, \cite{yu15lsun} leads to slightly more annotations but with low quality. For a fair comparison to manual annotation, we take into account the cost of the initial training set for all three verification approaches.

\subsection{Experiments on COCO and OpenImages}
\label{sec:simulationsCOCO}

\begin{figure}[t]
\center
\includegraphics[width=0.97\linewidth]{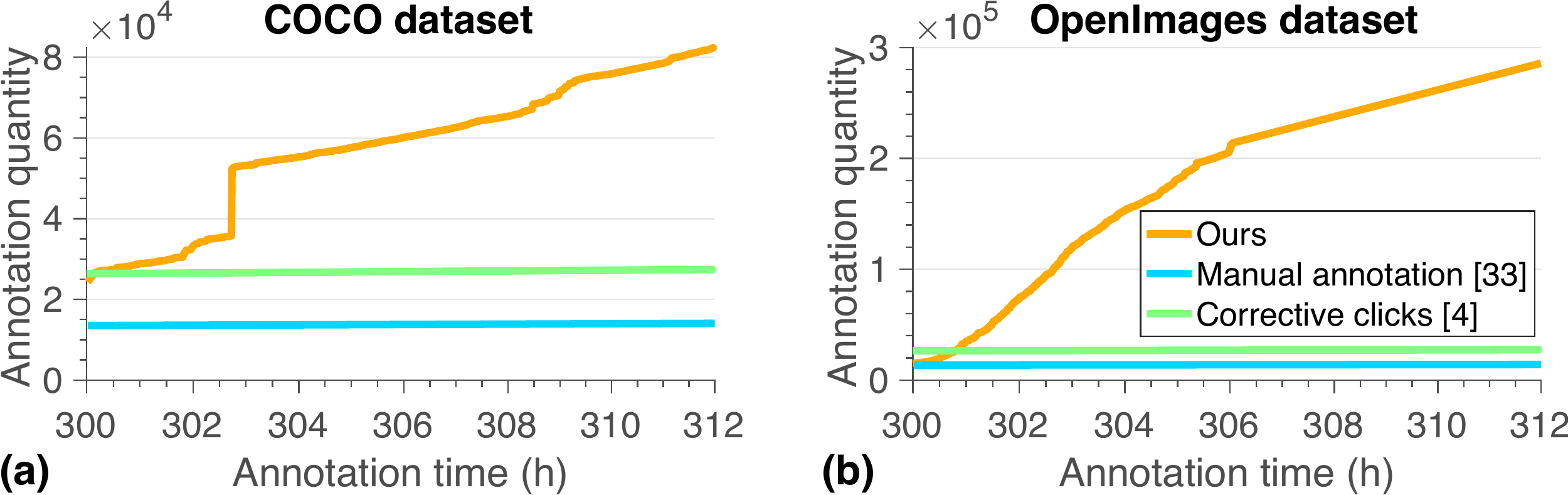}
\caption{\small \textbf{Experiments on (a) COCO and (b) OpenImages.} Annotation quantity vs. annotation time for~\cite{lin14eccv},~\cite{benenson19cvpr} and ours.}
\label{fig:experimentsCOCO_OID}
\end{figure}

Here we conduct simulated experiments on COCO~\cite{lin14eccv} and OpenImages~\cite{benenson19cvpr} (Fig.~\ref{fig:experimentsCOCO_OID}).
We follow the setting of Sec.~\ref{sec:simulationsADE} and \ref{sec:simulationsPLACES} exactly without tuning any parameters 
and train our initial segmentation model on 2k COCO images (80 classes). We run our pipeline on two unlabeled sets of images:
(a) the remaining 121k COCO images, and
(b) the OpenImages subset with a GT in the 60 COCO classes that overlap with the 350 OpenImages ones (316k images).

\mypar{Quantity and annotation time.}
We obtain (a) 86k COCO and (b) 270k OpenImages annotations with only 12 h (+300 h to annotate the initial set). 
With 312 annotation hours, we obtain 14k instances with~\cite{lin14eccv} and 28k with~\cite{benenson19cvpr} in either dataset (the two lines in Fig.~\ref{fig:experimentsCOCO_OID} appear horizontal because we manually annotate only 540 masks in 12 h~\cite{lin14eccv}).
Our time saving on OpenImages is 19$\times$ with respect to~\cite{lin14eccv}.

\mypar{Quality.}
The quality of our masks is 84.6\% on COCO. \cite{benenson19cvpr} reports 82.0\% for~\cite{lin14eccv} and 84.0\% for~\cite{benenson19cvpr} on a subset of COCO annotated with free-painting annotations.
On OpenImages, we obtain 85.0\% on the validation and test set vs 86.0\% with~\cite{benenson19cvpr} (provided by the authors of~\cite{benenson19cvpr}).

\mypar{Hyperparameters.} 
The annotation gain of our propagation method on COCO and OpenImages with the same hyperparameters tuned on ADE20K indicate that our method is not sensitive to per-dataset hyperparameter tuning.

\section{Large scale experiment on Places}
\label{sec:experimentPlaces}

In this section, we present our results on a large-scale experiment to obtain object segmentation masks using our framework with a fixed annotation budget.

\mypar{Data.} 
We use 1M images from the trainval set of the Places dataset~\cite{zhou14nips} as our unlabeled set and using our method we populate it with high-quality masks for 80 object classes. Unlike the simulated experiments (Sec.~\ref{sec:simulationsADE}) where the unlabeled pool is limited, using a pool of 1M images allows us to demonstrate the scaling of the annotation process and the gain of our clustering approach at a large scale.

\begin{figure*}[t!]
\center
\includegraphics[width=\linewidth]{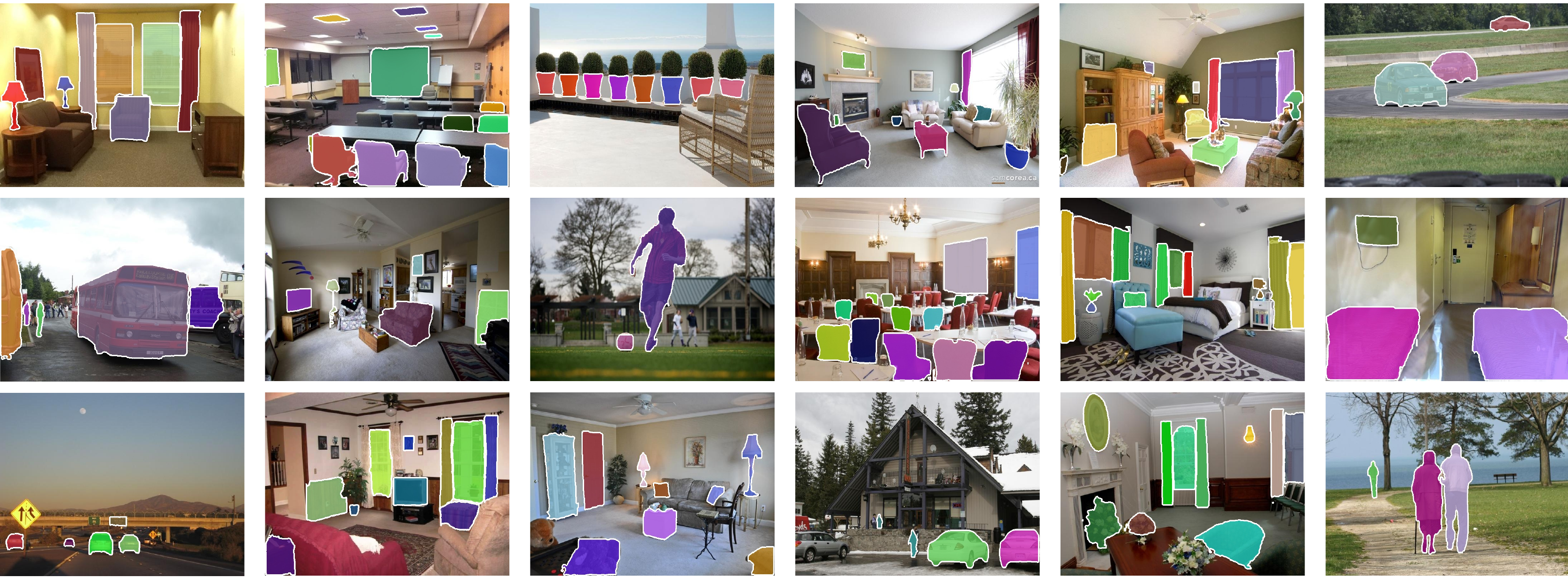}
\caption{\small \textbf{Obtained mask annotations in Places} using our proposed interactive approach under a small fixed annotation budget.}
\label{fig:qualReultsPlaces}
\end{figure*}

\mypar{Crowd-sourcing interface.}
To ensure high quality responses, we carefully design a crowd-sourcing protocol, while following common quality control  mechanisms~\cite{endres10cvprw,kuznetsova18arxiv,papadopoulos17iccv,papadopoulos17cvpr,russakovsky15ijcv,sorokin08cvprw,vondrick13ijcv,zhou17pami}. First, we provide a set of instructions with examples; then, the annotators must pass a qualification test to proceed to the annotation stage; and finally, we monitor performance by using hidden quality control images.
The annotators are shown a mask and a class label. They should respond positively if the mask outlines the object correctly ($\textrm{IoU} \ge 0.75$), and negatively otherwise. 

\mypar{Data collection.}
We train our segmentation model on 1,000 fully annotated images from the ADE20K training set~\cite{zhou17cvpr} and obtain more than 10M mask predictions on the 1M unlabeled Places. 
Then, we run our framework using all design choices and hyper-parameter values from Sec.~\ref{sec:method}-\ref{sec:simulations}.
We run in total 191,929 verification questions on Amazon Mechanical Turk, resulting in 730 high quality clusters ($\Tilde{Q}\ge0.85$). We accept in total 993,677
high quality object segmentation masks, which form our annotated dataset (Fig.~\ref{fig:qualReultsPlaces}).
The set of our annotations is 101$\times$ larger than our initial training set and 2.3$\times$ larger than ADE20K~\cite{zhou17cvpr}.


\mypar{Annotation time.}
The mean response time of the annotators is 1.4 s per binary question. The total time of the verification including quality control is 73 hours. Adding this time to the overhead for segmenting the initial set results in 290 annotation hours or less than \$3,000.
Note that manually drawing 1M polygons would require 22k hours and cost over \$200k~\cite{lin14eccv}, while the interactive segmentation of OpenImages~\cite{benenson19cvpr} would require about 11k hours. 
Our framework leads to a 76$\times$ speed up in time compared to~\cite{lin14eccv}.

\begin{figure}[t!]
\center
\includegraphics[width=\linewidth]{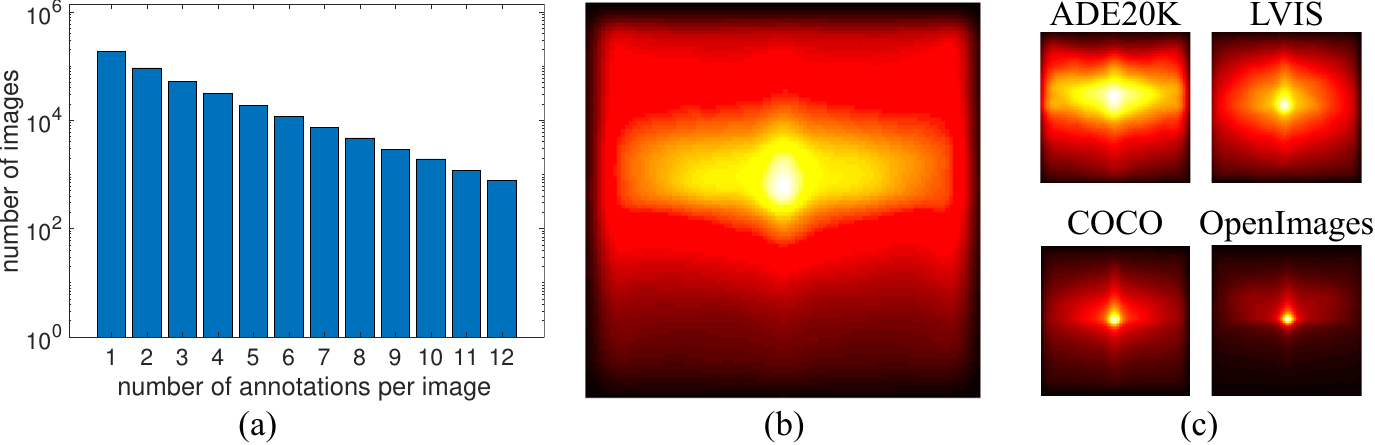}
\caption{\small \textbf{Statistics on Places annotations.} (a) \textbf{Density:} The number of our annotations per image. \textbf{Diversity:} The distribution of (b) our annotation centers compared to (c) ADE20K, LVIS, COCO and OpenImages V6 train sets.}
\label{fig:dataset_analysis}
\end{figure}

\mypar{Quality.}
To evaluate the segmentation quality $SQ$ of our masks, we randomly select 1,142 images from the unlabeled Places pool and an expert annotator manually draws accurate polygons around each instance. 
We achieve an $SQ$ of 81.4\% computed with 1,109 matching GT annotations. 
The quality of our masks is on par with that of other datasets:  82.6\% in COCO~\cite{kirillovOverview}, 83.9\% in Cityscapes~\cite{kirillov19cvpr}, 77.9\% in Vistas~\cite{kirillov19cvpr}, 84.0\% in OpenImages ~\cite{benenson19cvpr}, 85.0\% in LVIS~\cite{gupta19cvpr} and 85.9\% in ADE20K~\cite{kirillov19cvpr}.
Note that we use these numbers as a reference and not for straight comparison because the size of the instances, the number of classes and the scene complexity vary a lot across datasets. In Sec.~\ref{sec:simulationsCOCO}, we present a more fair comparison to~\cite{lin14eccv} and \cite{benenson19cvpr}.

\mypar{Annotation usefulness.}
We evaluate the practical value of the obtained annotations by training PointRend~\cite{kirillov20cvpr} on 1M masks (Places masks and about 10k initial masks). Training a vanilla model from sparsely annotated images can be catastrophic as proposals are sampled from unlabeled image regions and used as negatives for the classifier~\cite{niitani19cvpr,wang20arxiv,yoon20arxiv}. We do two simple modifications to avoid this. First, we lower the proposals batch size so that the fraction of positives and negatives is the same as in a fully annotated image. Second, we use the predicted non-accepted masks to avoid sampling negative proposals that highly overlap with them.
We achieve 12.7\% AP on the ADE20K validation set. As a reference, training on the initial fully annotated set achieves 8.1\% AP. Therefore, at a modest extra annotation time of 73 h we obtain a much better model. From another perspective, spending 73 h to manually annotate more objects and train
a model on 1,400 manually annotated images achieves 9.6\% AP. This shows that given the same annotation time, our method leads to a better model than manual annotation.
%

\mypar{Annotation coverage.}
We obtain an annotation coverage (fraction of instances found) of 9.7\% in Places (estimated on the annotated 1,142 images). A higher budget could lead to more masks since we did not fully annotate the trees. As a reference, we obtain a coverage of 27.1\% in ADE20K.


\begin{figure}[t!]
\center
\includegraphics[width=\linewidth]{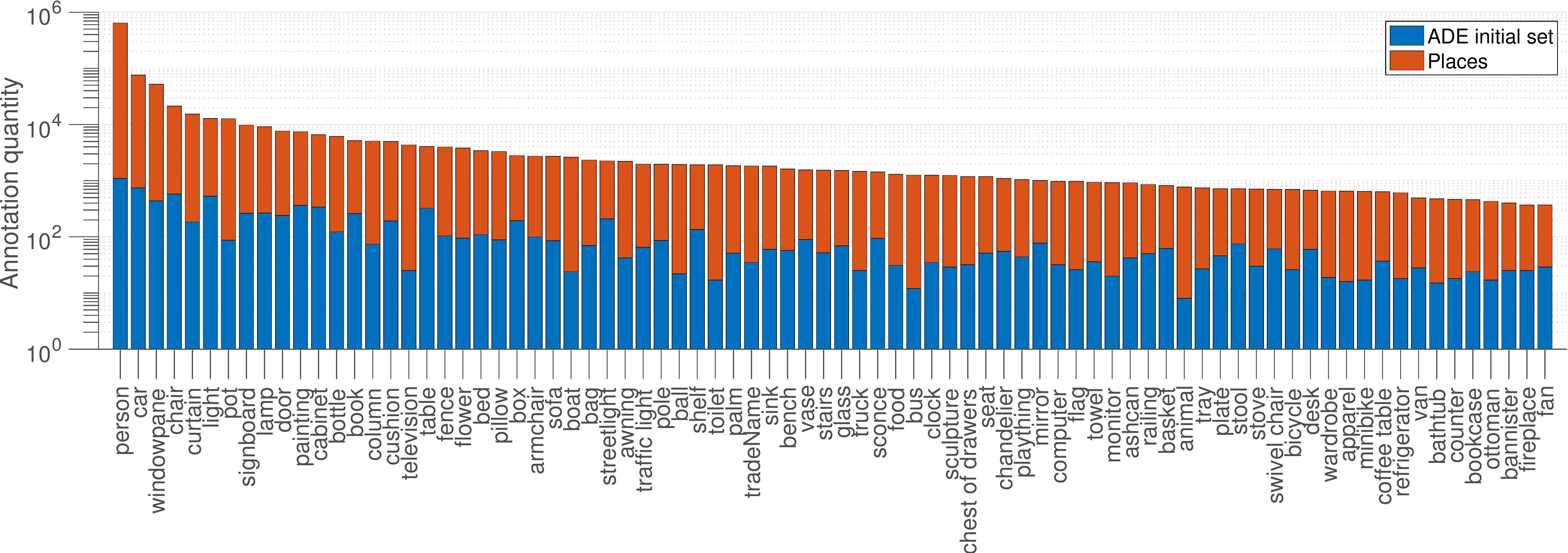}  
\caption{\small \textbf{Object class distribution.} The number of annotations for the initial set (blue) and the obtained Places annotations (red).}
\label{fig:dataset_distribution}
\end{figure}

\mypar{Density and diversity.}
Fig.~\ref{fig:dataset_analysis}(a) shows the histogram with the number of annotations per image. 
%
Fig.~\ref{fig:dataset_analysis}(b) shows the spatial distribution of our obtained mask centers in normalized image coordinates. 
Even though none of our masks were drawn manually, we observe a quite diverse spatial distribution of complexity.
Fig.~\ref{fig:dataset_distribution} shows the object class distribution of our annotations compared to the initial training set. With an extra time of only 73h, we obtain two orders of magnitude more annotations for most of the classes.

\section{Conclusions}

We presented a highly efficient pipeline for annotating object segmentation masks. 
Our pipeline overpasses the major limitation of manual or interactive annotation where the cost grows linearly with the number of annotations. 
Instead, it exploits the commonalities between objects at a large scale and quickly propagates few verification labels to many examples. 
By applying our pipeline to a large-scale experiment, we obtained 1M masks with only 290 annotation hours. 
%
%
Our work marks a new direction for exploring the scaling of instance annotation. 
%
%
Future work involves turning the obtained sparsely labeled images into fully-annotated ones and annotating uncommon classes appearing in the long tail distribution as well as stuff classes.

\mypar{Acknowledgments.} This work is supported by the Mitsubishi Electric Research Laboratories. We thank V. Kalogeiton for helpful discussion and proofreading.

{\small
\bibliographystyle{ieee_fullname}
\bibliography{shortstrings.bib,dimBibTex.bib}
}

\end{document}